## Article

# Taking second-life batteries from exhausted to empowered using experiments, data analysis, and health estimation

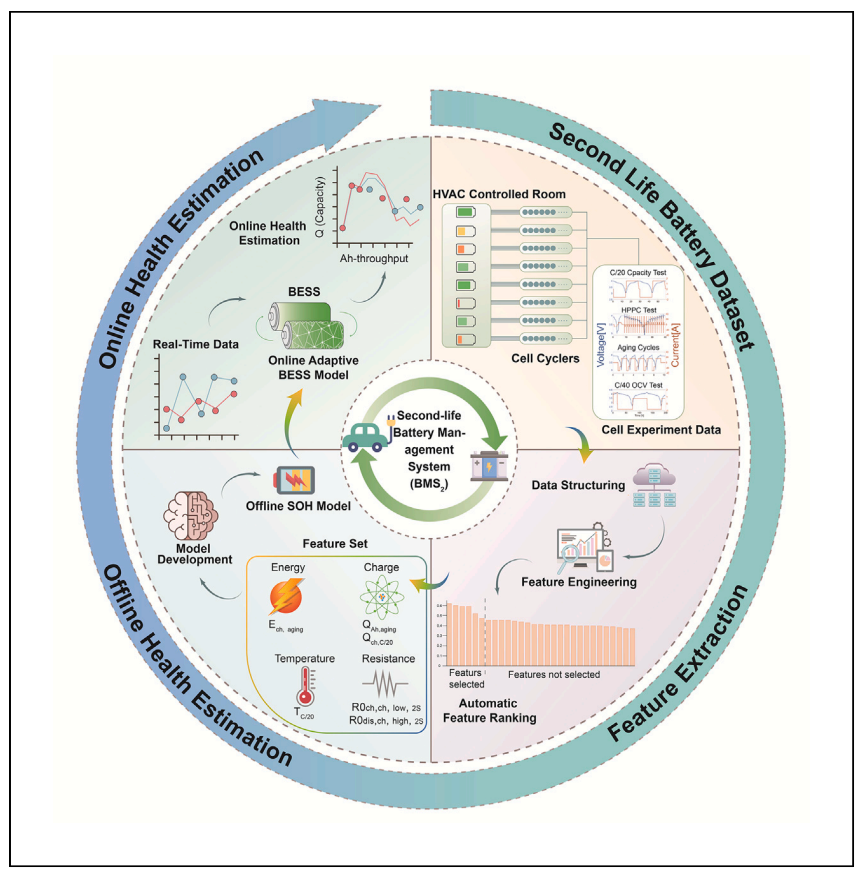

Xiaofan Cui, Muhammad Aadil Khan, Gabriele Pozzato, Surinder Singh, Ratnesh Sharma, Simona Onori

sonori@stanford.edu

### Highlights

Lab-collected, publicly available dataset from eight Nissan Leaf retired cells

Retired batteries may support the grid for a decade under certain conditions

Data-driven offline state-of-health estimation with online-accessible features

Online adaptive state-of-health estimation based on clustering to bound estimation errors

Here, Cui et al. introduce innovative offline and online health estimation methods for integration into a second-life battery management system for repurposed batteries in grid energy storage applications. Experimental data from retired electric vehicle batteries demonstrate that these batteries can reliably support the grid for over a decade.

## Article

# Taking second-life batteries from exhausted to empowered using experiments, data analysis, and health estimation

Xiaofan Cui,[1,3,5] Muhammad Aadil Khan,[1,5] Gabriele Pozzato,[1,4] Surinder Singh,[2] Ratnesh Sharma,[2] and Simona Onori[1,6,*]

## SUMMARY

The reuse of retired electric vehicle batteries in grid energy storage offers environmental and economic benefits. This study concentrates on health monitoring algorithms for retired batteries deployed in grid storage. Over 15 months of testing, we collect, analyze, and publicize a dataset of second-life batteries, implementing a cycling protocol simulating grid energy storage load profiles within a 3–4 V voltage window. Four machine-learning-based health estimation models, relying on online-accessible features and initial capacity, are compared, with the selected model achieving a mean absolute percentage error below 2.3% on test data. Additionally, an adaptive online health estimation algorithm is proposed by integrating a clustering-based method, thus limiting estimation errors during online deployment. These results showcase the feasibility of repurposing retired batteries for second-life applications. Based on obtained data and power demand, these second-life batteries exhibit potential for over a decade of grid energy storage use.

## INTRODUCTION

The global demand for lithium-ion batteries (LIBs) in grid battery energy storage systems (BESSs) is projected to exceed 500 GWh by the year 2030.[1] Simultaneously, over 200 GWh of electric vehicle (EV) batteries will reach the end of their first life (FL) by 2030.[2] These retired EV batteries are estimated to retain a significant portion (approximately 70%–80%) of their original energy capacity and power capability.[3] As a result, they can contribute up to a storage capacity of 140–160 GWh, which accounts for approximately 30% of the demand for grid energy storage services. Not only does the repurposing of EV batteries for grid BESSs offer economic benefits, but it also provides sustainability benefits.[4] From an economic standpoint, it can lead to a reduction in the manufacturing costs estimated to range between 157 to 526 $/kWh.[5] Furthermore, the second-life (SL) market has the potential to create a new revenue stream for retired EV batteries, allowing buyers to benefit from discounted upfront costs.[6] From a sustainability standpoint, the burden of $CO_2$ emissions associated with the production of new battery packs, estimated to be 123 $kgCO_2eq/kWh$, would be alleviated.[5,7]

For the purpose of this work, battery state of health (SOH) is defined as the ratio of the actual capacity to the initial (or rated) capacity.[8] Due to various usage conditions in their FLs, retired EV batteries inherently contain a wide spread in their SOH. Accurate, real-time SOH estimation is required to (1) ensure and enhance real-time operational safety, (2) enable proactive maintenance or replacement, (3) facilitate

[1]Department of Energy Science and Engineering, Stanford University, Stanford, CA 94305, USA

[2]Relyion Energy Inc., Santa Clara, CA 95054, USA

[3]Present address: Department of Electrical and Computer Engineering, University of California, Los Angeles (UCLA), Los Angeles, CA 90095, USA

[4]Present address: Form Energy, Berkeley, CA 94710, USA

[5]These authors contributed equally

[6]Lead contact

*Correspondence: sonori@stanford.edu

https://doi.org/10.1016/j.xcrp.2024.101941

optimized charging or discharging, and eventually (4) improve the lifetime as well as cost efficiency of batteries. Traditional battery management systems (BMSs) are tasked with cell balancing, thermal management, optimized charging, and communicating with the peripherals inside an EV for state-of-charge (SOC) and SOH monitoring. However, they are designed to work with fresh cells that generally have minimum variation in their internal parameters.

When retired batteries are repurposed for a new application, a new SL BMS ($BMS_2$) should be designed to suit the requirements of the new use case. Some key considerations in designing $BMS_2$ for repurposed batteries are (1) understanding the specific requirements of the new application. Different applications (e.g., stationary grid energy storage, EV charging, backup power) have unique demands, and the $BMS_2$ should be tailored accordingly. (2) Integrating safety features relevant to the new application. This may include overcurrent protection, temperature control, and fail-safes to ensure safe operation. (3) Developing balancing algorithms suitable for the repurposed batteries. (4) Defining charge and discharge control strategies based on the requirements of the new application to avoid stressing the SL batteries and ensure longevity. (5) This paper focuses on the design of health-monitoring techniques to be included in $BMS_2$.

Despite the extensive literature around LIB SOH estimation for conventional BMSs, practically no work exists on online SOH estimation for $BMS_2$. Existing studies in industry and academia on retired battery repurposing are primarily centered on the development of dedicated power electronics hardware and energy control systems.[9–13] For instance, ReJoule is focusing on developing a smart BMS chip that, if integrated into new vehicles, would enable real-time optimization of the battery system throughout its life cycle and thereby facilitate a smoother transition from FL-to-SL adoption.[9] Relectrify offers a combination of power electronics hardware and control software packages, granting granular control over individual SL battery cells or modules.[10] Elektroautomatik's technology focuses on the development of high-efficiency bidirectional programmable power supplies to interface the retired EV batteries with the alternating current (AC) grid,[11] and Smartville developed a refurbishing process for heterogeneous battery packs, which is based on customized power converters aimed at achieving SOH balance.[12] Lastly, a new hierarchical power processing architecture is developed to extract maximum performance from heterogeneous batteries with minimum power electronics cost.[13] To the best of our knowledge, these efforts have not adequately addressed the online health estimation problem of SL batteries once they are deployed in the field.

Current SOH estimation methods mainly include semi-empirical models, physics-based models, and data-driven models.[14] The use of these models is widespread for FL applications, but limited work exists for SL applications. Despite achieving good accuracy under specific laboratory conditions, semi-empirical models do not perform well when applied to unseen real-world scenarios[14] such as fluctuating temperatures and varying load profiles commonly seen in grid BESSs. Physics-based models are based on a set of partial differential algebraic equations and provide a wealth of information about electrochemical states; however, these models are limited in their ability to describe complex aging mechanisms inherent in SL batteries.[15] Except for the growth of the solid-electrolyte-interphase (SEI) layer and lithium plating, many other aging mechanisms, such as electrolyte decomposition and transition metal dissolution, are challenging to capture by physics-based models. Furthermore, retired battery cells, even from the same EV, aged differently during their FL. Despite the existing attempts to utilize physics-based models for real-time battery state estimation,[16] these models are computationally intensive and seldom implemented for online SOH estimation in real BMSs.

Data-driven SOH estimation methods have gained considerable attention due to their ability to surrogate multiple interlaced aging mechanisms and handle various aging conditions without the need to model them explicitly.[17,18] Machine learning (ML) techniques for SOH estimation include artificial neural networks,[19–21] linear regression,[20] support vector regression (SVR),[22] Gaussian process regression (GPR) with bagging, etc.[23] However, for SL batteries, data-driven SOH estimation remains a challenging task. Existing works around SOH estimation for SL batteries have used publicly available FL datasets,[19,24] while other works performed analysis on laboratory-aged SL batteries.[15,25] Actual SL batteries from 94 battery packs retired from a Chinese EV bus,[21] six lithium iron phosphate battery cells retired from a Chinese electric passenger car,[20] and 32 modules retired from Nissan Leaf EVs[26] have also been used. However, these works collected aging data under nearly constant temperature conditions, and the datasets were not made publicly available. On the contrary, in a grid BESS composed of multiple battery packs, battery temperature fluctuates for several reasons: (1) a practical cooling system for grid BESSs cannot completely reject the temperature disturbance caused by ambient temperature variation[27] and (2) uneven cooling creates intra-cell temperature variations inside the pack.[28]

The performance of data-driven models is highly dependent on the input features selected to train these models, which in turn depend on the type of excitations used. Features are generally selected to be highly correlated to the target output, e.g., battery capacity, and existing works have reported high SOH correlation features from entire charging curves,[29] constant-current (CC) voltage curves and constant-voltage (CV) current curves,[23] incremental capacity (IC) curves,[20,30] and voltage curves.[31] However, from a feature selection perspective, most of these features are obtained from reference performance tests (RPTs) that cannot be conducted once the batteries are deployed in the field. This indicates that SOH estimation models need to utilize features that are accessible onboard in order to provide continuous health monitoring for SL batteries.

Adaptive SOH estimation for FL batteries has used model-based observers[16] leveraging physics-based models, but this can be computationally intensive for on-board scenarios. Other approaches based on data-driven models update model weights using feedback of periodic SOH measurements through RPTs.[32–34] The measured incoming current, voltage, and temperature data for real grid applications are prone to uncertainties. Power demand from the grid determines the current and voltage to BESS, and daily temperature variations affect the onboard measured temperature. If the ML SOH estimation algorithm were to be deployed, the unseen and stochastic nature of these signals could lead to unbounded errors in health estimation. To overcome these issues without disrupting BESS operation, an adaptive SOH estimation algorithm is required in $BMS_2$ applications that ensures the estimation error remains bounded onboard the BMS.

The contributions of this work are as follows.

(1) First SL battery aging dataset made publicly available: aging and RPT data were collected from eight retired lithium manganese oxide (LMO)/graphite pouch cells from Nissan Leaf battery packs over 15 months. Refer to the data and code availability section for more information.

(2) Aging protocol engineered to cycle retired batteries for grid energy storage: SL battery aging is conducted based on a simplified grid duty cycle protocol that replicates a peak shaving scenario with voltage derating to cycle between 3 and 4 V.

(3) New SOH estimation methods for $BMS_2$:

- Offline SOH estimation: a comprehensive feature extraction and selection process is carried out to identify features that are highly correlated to SOH and also available online for $BMS_2$. The performance of four offline ML-based SOH estimation models, with selected features as input and cell capacity as output, is compared using significance testing.
- Adaptive online SOH estimation: a clustering-based SOH estimation model is developed that works in conjunction with the offline SOH model to guarantee bounded estimation error for unseen, real-time incoming data. The method provides a way to ensure safe and long-lasting operation of SL batteries.

## RESULTS AND DISCUSSION

### SL battery aging dataset

The dataset used in this study is from eight pouch cells extracted from two distinct retired EV battery packs with LMO/graphite chemistry. The fresh cells have a nominal capacity of 32.5 Ah with minimum/maximum voltage limits of 2.5–4.2 V. Given the unknown history of these cells, an initial RPT was conducted to evaluate their residual capacity. Subsequently, an aging campaign was carried out to investigate the degradation behavior of these cells, as illustrated in Figure 1. Three unique characteristics of this dataset include the following:

(1) RPTs conducted under varying temperatures: during a 15-month-long experimental campaign, the cells are tested in a laboratory environment where the temperature was regulated by a heating, ventilation, and cooling (HVAC) system based upon human comfort and seasonal variations. The measured cell surface temperature obtained from the aging cycles portion of the campaign varied between 20°C and 35°C. Across the tested cells, the RPT suite includes the C/20 capacity test, the hybrid pulse power characterization (HPPC) test, and the C/40 open-circuit voltage test, as shown in Figures 1A, 1B, and 1C, respectively. Furthermore, measured temperature data obtained from C/20 capacity tests showed variation between 17°C and 29°C. Please refer to Note S2 for further details about the aging campaign.

(2) Aging cycles mimic load profiles observed in grid energy storage used for peak shaving applications: aging of SL batteries in the experimental campaign is based on an engineered current protocol that replicates a grid storage duty cycle.[35] As shown in Figure 1D, discharge current corresponds to night-time BESS usage, while the charge portion corresponds to daytime usage. The 1C CC discharge reflects high battery usage during late evening and early night hours. Once the SOC reaches 50% (100% SOC is defined by an upper cutoff voltage of 4 V, and 0% SOC is defined by a lower cutoff voltage of 3 V), the C-rate changes to C/2, corresponding to low battery usage during the remainder of the night until 0% SOC (or 3 V). Afterward, the battery is CC charged at a C/2 C-rate until it reaches 100% SOC (or 4 V). At this point, the battery is said to have completed one single cycle of operation.

(3) Modified cutoff voltages: cutoff voltages are usually specified by the manufacturer, and it is common to charge/discharge cells between their maximum and minimum voltage range. Voltage derating has been shown to decelerate the degradation of new batteries,[36–38] leading to a decrease in the rate of loss of

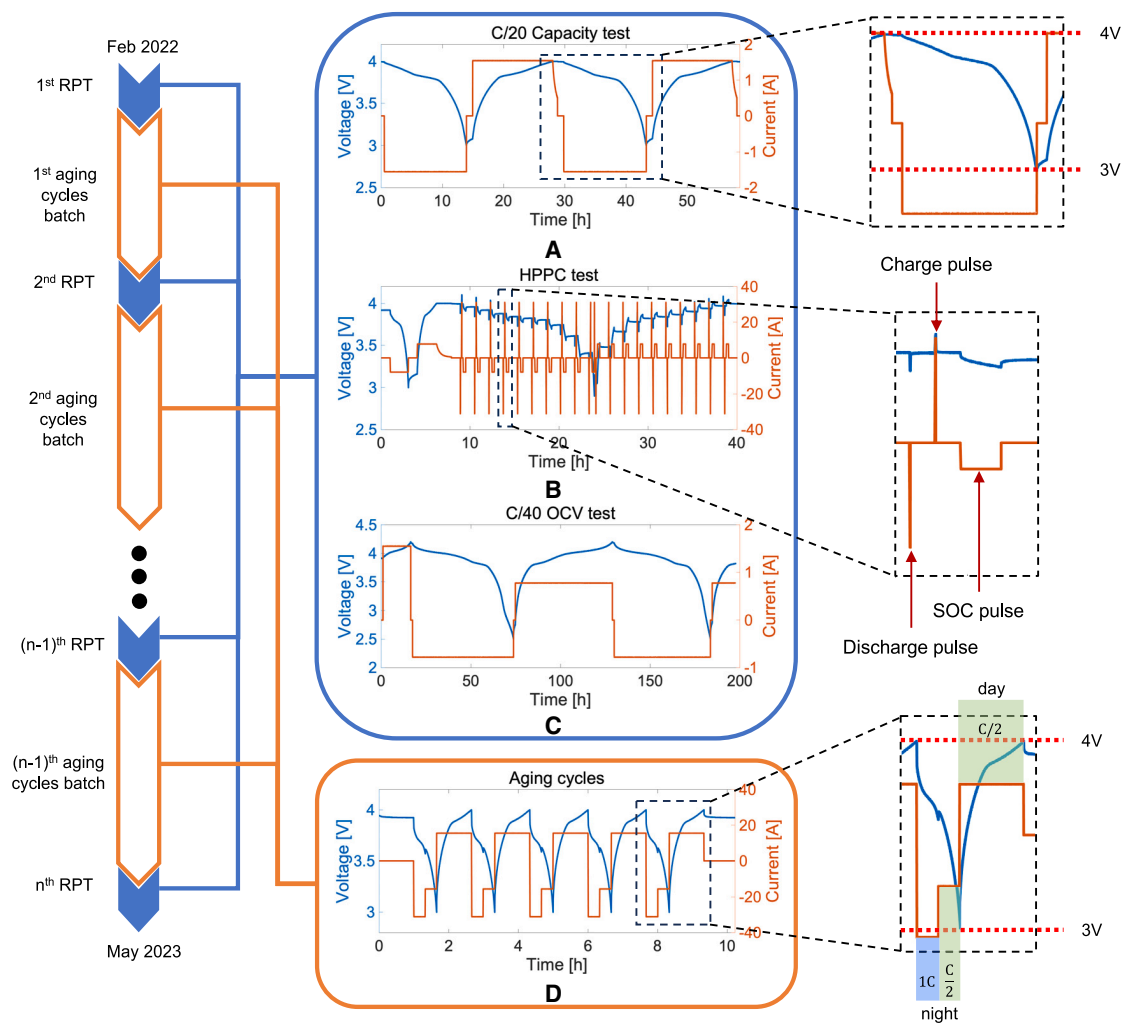

**Figure 1. Schematic of the experimental campaign conducted on eight pouch cells (named cells 1.1, 1.2, 1.3, 1.4, 2.1, 2.2, 2.3, and 2.4) tested in this work**

(A–C) Battery cells are characterized through RPTs, namely, (A) C/20 capacity test over 3–4 V voltage window, (B) HPPC test executed in both discharge (from 4 to 3 V) and charge (from 3 to 4 V), and (C) C/40 open-circuit voltage (OCV) test executed from 4.2 to 2.5 V.

(D) Following the 1st RPT, cells are cycled between 3 and 4 V (according to the synthetic aging cycle) followed by the 2nd RPT, and so on. The data analyzed in this work were collected from February 2022 to May 2023.

capacity. In the dataset collected and analyzed in this work, the C/20 capacity tests, HPPC tests, and aging cycles were performed within a modified cutoff voltage range between 3 and 4 V. The C/40 capacity test is the sole exception, conducted within the maximum and minimum cutoff voltage of 4.2 and 2.5 V.

## Analysis of experimental data

The experimental data collected during the aging campaign go through the pipeline given in Figure S2. To understand how the data are structured, the reader is referred to Note S3 (data structuring section). The C/20 charge capacity $Q_{ch,C/20}$ is shown as a function of Ah throughput in Figure 2A. The tested cells manifest similar charge and discharge capacity trends, which is why $Q_{ch,C/20}$ is selected as the SOH indicator as well as the output of the health estimation model. The eight cells have different initial C/20 charge capacities between 17 and 25 Ah, and they starts to increase with Ah throughput during the initial few months of the aging campaign. It is interesting to note that over a period of 15 months, despite experiencing approximately 100,000 Ah throughput, these batteries saw an increase of $Q_{ch,C/20}$ between 0.6% and 16.1% by the end of the campaign. This kind of capacity behavior is unconventional and rarely reported in literature. Through an operational lifetime analysis (Note S1), it is shown that these SL batteries have the potential to last for 14 to 18 years in both commercial and residential BESS applications[39] under specific derated conditions and HVAC-regulated operational temperatures.

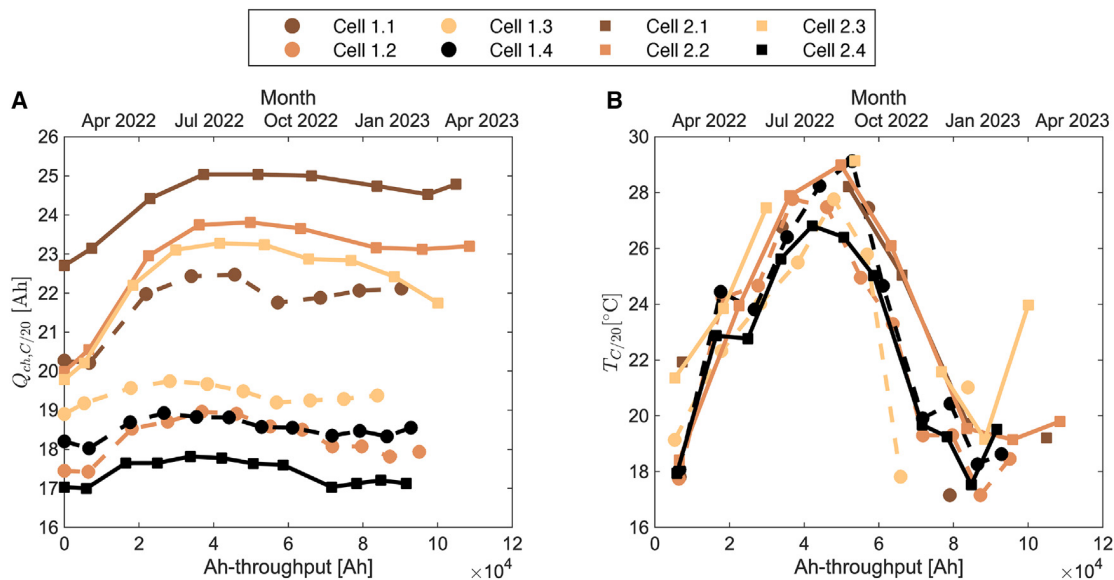

**Figure 2. C/20 capacity and temperature variation as a function of Ah throughput and months of the year for eight SL cells in this work**
(A) Variation of C/20 charge capacity $Q_{ch,C/20}$.
(B) Variation of temperature $T_{C/20}$ during C/20 capacity tests from February 2022 to April 2023.

Voltage derating has been shown to reduce the rate of degradation,[36] but an increase in capacity might not be attributed only to this operation. The observed increase in capacity of the tested SL batteries is postulated to be dependent on the seasonal temperature variations over the testing period. As shown in Figure 2B, the temperature $T_{C/20}$, which is measured during C/20 capacity tests only, rises during the middle of the year and reaches its peak in August/September 2022 before decreasing again. As $T_{C/20}$ increases, the capacity of the cells also increases, which could be explained with an increase of the chemical reaction rates inside the cells leading to a decrease in charge-transfer resistance.[40] Lower charge-transfer resistance suggests more Li ions cross the SEI layer and intercalate into the electrode, which increases the observed $Q_{ch,C/20}$ of the cells. A relatively decreasing capacity trend as temperature decreases indicates that temperature plays a dominant role in the variation of cell capacity as compared to other factors, such as the number of aging cycles.

### Analysis of selected features

In this paper, a comprehensive feature extraction process was carried out with the aim of extracting features from real-time operation data. A total of 66 features were initially extracted from RPTs, aging cycles, and temperature data (see Table S2) consisting of IC peaks, aging cycle resistance, and more. With these features in hand, an automatic feature selection method was used to identify a subset of features that were highly correlated to SOH (see experimental procedures). As shown in Figure 7, from the maximum relevance-minimum redundancy (mRMR) algorithm, a set of six features that are ranked the highest is selected.

**Table 1. Initial C/20 charge capacity $Q_{initial,ch,C/20}$ for the eight cells tested in this work**

| Cells | 1.1 | 1.2 | 1.3 | 1.4 | 2.1 | 2.2 | 2.3 | 2.4 |
|---|---|---|---|---|---|---|---|---|
| $Q_{initial,ch,C/20}$ (Ah) | 20.27 | 17.42 | 19.57 | 18.93 | 25.04 | 23.65 | 22.87 | 17.59 |

The first selected feature is the initial C/20 charge capacity $Q_{initial,ch,C/20}$, whose value across the eight cells is shown in Table 1, while the other five features are shown in Figure 3. In Figure 3A, accumulated aging cycle Ah-throughput $Q_{Ah,aging}$ varies linearly with aging cycles for all the cells. Figure 3B shows aging cycle charge energy-throughput $E_{ch,aging}$, which is only calculated during the charging period of aging cycle profile. For reference, $Q_{ch,C/20}$ is also shown in this figure, and it displays the same trend as $E_{ch,aging}$. Generally, it is recognized that resistance exhibits behavior opposite to that of capacity, as illustrated by the two resistance features in Figure 3C. Specifically, the 2 s charge pulse resistance from charge HPPC, denoted as $R0_{ch,ch,low,2s}$, and the 2 s discharge pulse resistance from charge HPPC, denoted as $R0_{dis,ch,high,2s}$, both exhibit a decrease followed by an increase for the majority of cells. It is noted that some measurement errors during a couple of HPPC tests resulted in anomalous resistance behavior for some of the cells, e.g., cell 2.3, in between 1,000 and 2,000 aging cycles. Finally, the last feature $T_{aging}$, as shown in Figure 3D, shows similar behavior to $T_{C/20}$ in Figure 2B, which highlights that the influence of seasonal variation of temperature was present consistently throughout the experimental campaign. $T_{aging}$ reaches a higher peak value of around 35°C because it captures the behavior of temperature at a higher resolution due to the large number of aging cycles as compared to the number of RPT tests. This suggests that the cells experienced temperatures of greater than 30°C for a significant number of aging cycles. Further details about the features and their naming conventions are provided in Note S3.

The SOH estimation model is designed with the ultimate goal to run onboard BMS₂, so the features it relies on are purposely selected to be available online for the model. The value of $Q_{initial,ch,C/20}$ is derived from the C/20 capacity test. It is reasonable to assume that, prior to repurposing retired batteries into an SL energy storage system, an initial capacity assessment is conducted, and the resulting information can be fed into the model. $Q_{Ah,aging}$ and $E_{ch,aging}$ are obtained from aging cycles, directly from current and voltage during operation. Furthermore, $T_{aging}$ is simply the measured temperature during battery operation, which is reasonably assumed to be available through onboard temperature sensors. Lastly, although the resistance features $R0_{ch,ch,low,2s}$ and $R0_{dis,ch,high,2s}$ are obtained from HPPC test, power converters in BMS₂ should be capable of generating high-frequency, short-duration pulses without adversely affecting the health of the cells.[41] Based on this, it can be concluded that all six features are available onboard, allowing the use of updated battery information for reliable SOH estimation.

### Offline SOH estimation using data-driven approach

Data-driven SOH estimation approaches are generally divided into two categories: supervised learning and unsupervised learning. In this work, we use supervised learning methods for the health estimation task by utilizing the six selected features, discussed in the previous section, as the inputs and SOH as the output of the model. Four distinct data-driven models, namely, elastic net regression (ENR), SVR, random forest regression (RFR), and GPR are trained on six cells and tested on two cells. Details about data preprocessing and model descriptions are given in Note S3. Model hyperparameters are optimized using either grid search or Bayesian optimization along with 5-fold cross-validation. The performance metrics (see experimental procedures) are given in Table 2, which shows that the ENR model performs the best on

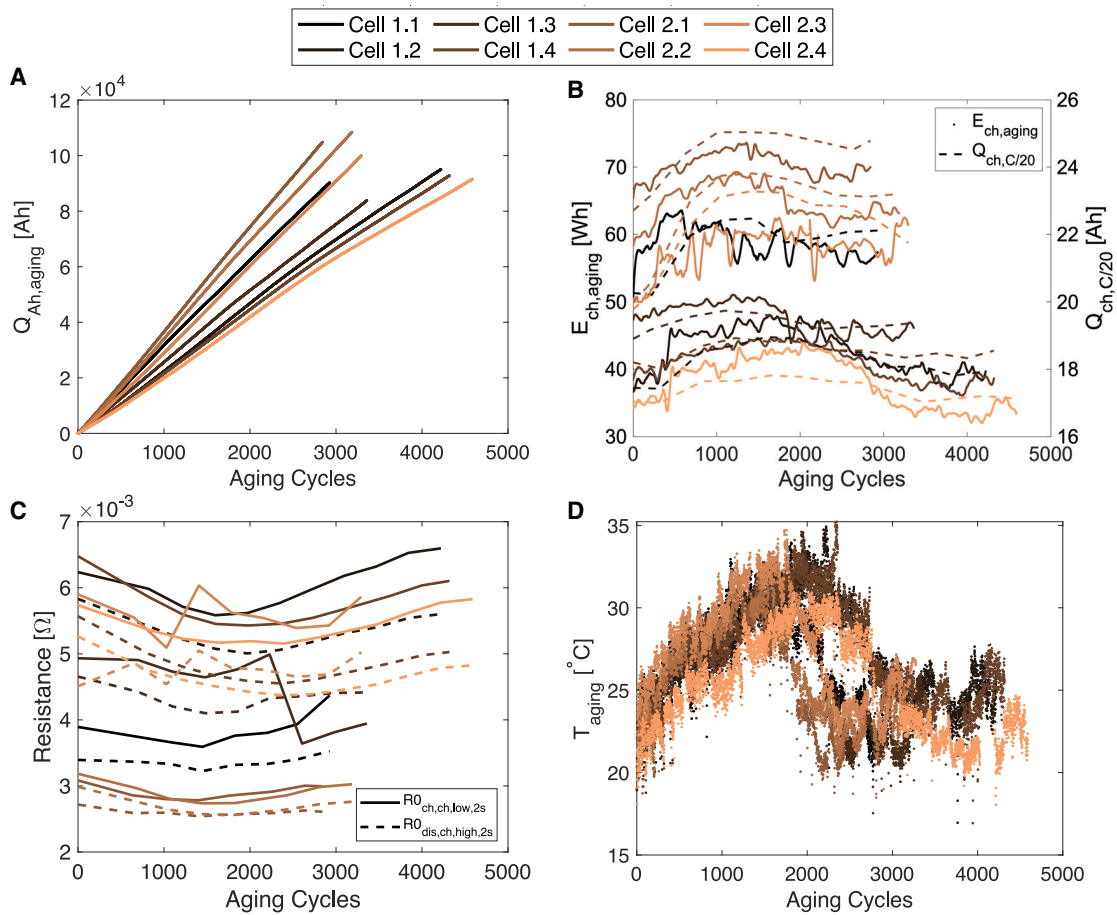

| Cell 1.1 | Cell 1.3 | Cell 2.1 | Cell 2.3 |
| Cell 1.2 | Cell 1.4 | Cell 2.2 | Cell 2.4 |

**Figure 3. Five selected features from automatic feature selection**

(A) Aging cycle accumulated Ah-throughput $Q_{Ah,aging}$.

(B) Aging cycle charge energy-throughput $E_{ch,aging}$ is the energy calculated during the charging events of the aging cycles, and it shows a similar trend as $Q_{ch,C/20}$ overplotted in the same figure.

(C) Resistance features $R0_{ch,ch,low,2s}$ and $R0_{dis,ch,high,2s}$.

(D) Average aging cycle temperature $T_{aging}$ showing a similar increase followed by a decrease to $T_{C/20}$ in Figure 2B. All features are shown with respect to aging cycles.

the test set with a root-mean-squared percentage error (RMSPE) of 1.8% and a mean absolute percentage error (MAPE) of 1.38%. For this test set, other models have an RMSPE and a MAPE of greater than 3% despite the better performances of RFR and GPR on training data. This indicates that the GPR and RFR models might suffer from overfitting, which leads to poor generalization. To further validate the performance of these models, they are tested on eight different test sets, and the results are given in Figure S7. Through significance testing (Note S4), it is observed that the ENR model performs better than the SVR and RFR models, but the GPR model has a comparable performance, albeit slightly worse. Based on these results, the ENR model is chosen for the remainder of this work.

In Figure 4, the ENR model shows good estimation performance on the test set, which includes cell 1.4 and cell 2.4. Furthermore, the presence of small errors in the estimation results of the training cells indicate that the model learns from the training data, but it does not overfit to them, most likely due to the presence of $L_1$ and $L_2$ regularization in the model. Overfitting to the training data would give perfect estimation results on all the training cells but large errors on the test cells. Figure 5 shows the distribution of pointwise capacity estimation percentage error (PCEPE) for the training and test cells shown in Figure 4. The errors are distributed around zero, with the majority of the errors within the 10th and 90th percentiles for both the training and test sets. The robustness of the ENR model is shown in Figure S8 by training and testing the model on different combinations of cells. As shown in Figure S9, the MAPE is 2.3% or less for three out of four test sets. Despite the size of the dataset, regularization in the ENR model and highly correlated input features help the model to perform well across a range of test sets. Furthermore, it is computationally inexpensive to train this ENR model, and the size of the trained model is in the order of kilobytes (KB) in MATLAB.

**Table 2. Comparison of SOH estimation performance, in terms of RMSE, RMSPE, and MAPE, for four different machine learning models developed in this work**

|  | Cells | Metrics | ENR | SVR | RFR | GPR |
|---|---|---|---|---|---|---|
| Training set | 1.1, 1.2 | RMSE (Ah) | 0.51 | 0.85 | 0.27 | 0.37 |
|  | 1.3, 2.1 | RMSPE (%) | 2.34 | 3.99 | 1.26 | 1.71 |
|  | 2.2, 2.3 | MAPE (%) | 1.88 | 1.68 | 0.98 | 1.12 |
| Test set | 1.4, 2.4 | RMSE (Ah) | 0.32 | 0.82 | 0.90 | 0.61 |
|  |  | RMSPE (%) | 1.80 | 4.59 | 5.17 | 3.42 |
|  |  | MAPE (%) | 1.38 | 3.26 | 4.46 | 3.03 |

**Adaptive SOH estimation using online-available features**
Batteries used for grid applications experience a lot of variations in their power demand, which means that the incoming current, voltage, and temperature data can have uncertainty. Direct deployment of the ENR model on $BMS_2$ is not robust and does not guarantee any boundedness on the estimation error, which can cause the SOH estimation to diverge significantly, resulting in unreliable estimates. To avoid this, an adaptive estimation algorithm is proposed in this work that uses clustering-based estimation to bound the SOH estimation error from the ENR model[42] (algorithm given in Note S5). Figure 6

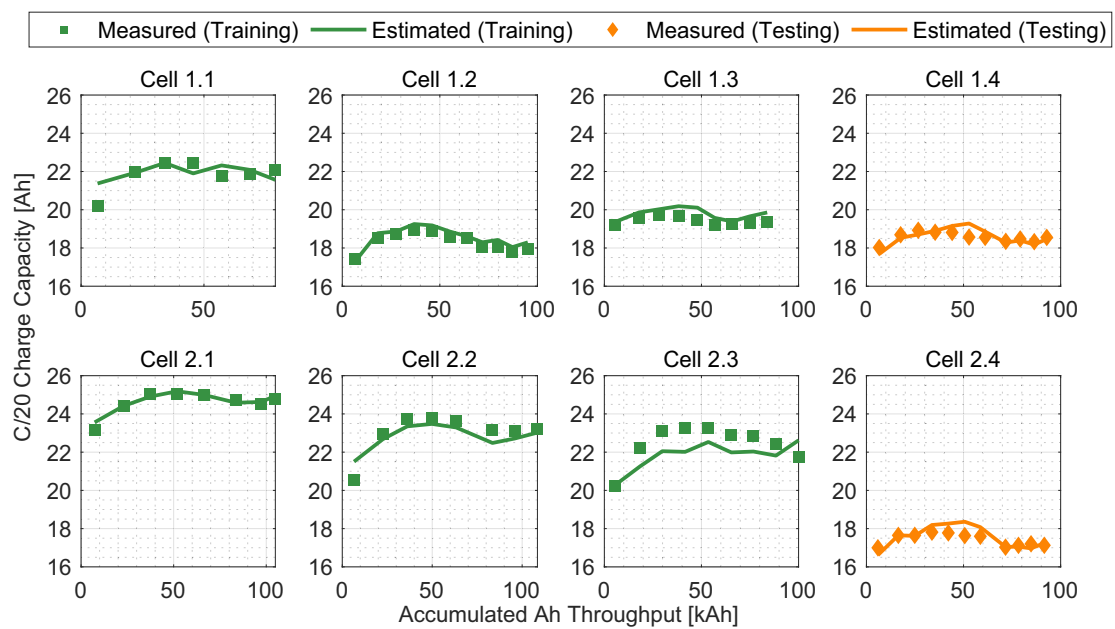

**Figure 4. Comparison between the measured and estimated battery capacity for each individual cell**
The C/20 charge capacity of each cell is estimated using the ENR model. The training set comprises cells 1.1, 1.2, 1.3, 2.1, 2.2, and 2.3, while the test set includes cells 1.4 and 2.4. The x axis represents accumulated Ah throughput, which monotonically increases with time as cells undergo cycling.

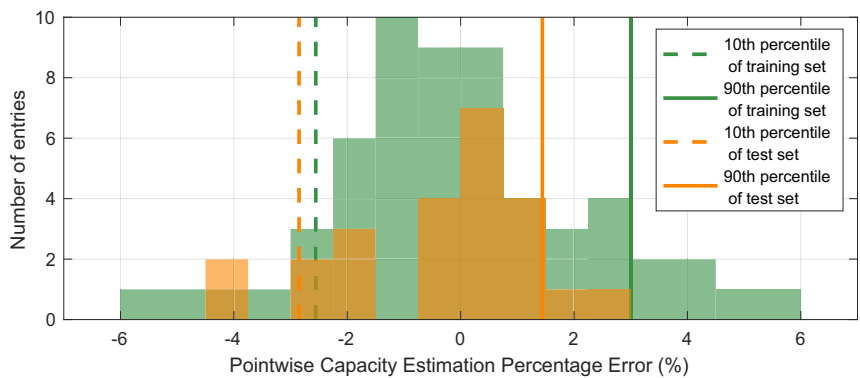

**Figure 5. Pointwise capacity estimation percentage error (PCEPE) histogram of the offline capacity estimation**

PCEPEs of Figure 4 are presented in the histogram. The PCEPEs, defined as the absolute difference between the estimated and true SOH, normalized by the true SOH, are distributed around zero on both training and test sets. The 10th and 90th percentiles of the PCEPEs on the training set are −2.55% and 3.01%, respectively, as shown in the green vertical lines. The 10th and 90th percentiles of the PCEPEs on the test set are −2.85% and 1.44%, respectively, as shown in the orange vertical lines.

shows the performance of the adaptive estimation model on cell 2.4, with the remaining cells in the training set. It can be observed that the estimations of the offline ENR and adaptive models coincide in the beginning (at low Ah throughput), but as more time series data become available, the adaptive model incorporates information from the complete trajectory. For cell 2.4, the maximum PCEPE is 2.7% for the offline ENR, but it reduces to 1.55% for the adaptive model.

Table S4 shows the comparison of RMSPEs between the adaptive model and the offline ENR model for all eight test cells. Due to the absence of true $Q_{ch,C/20}$ data in real applications, the estimation from the offline ENR model is the benchmark. The adaptive model is developed to ensure estimation errors remain bounded, while the improvement in estimation errors is an added possibility based on the closeness in feature space of the input features of the training and test cells. Although more than half of the test cells show a decrease in RMSPE with adaptive estimation, some cells have an increased RMSPE, such as cell 2.1. However, even though only $Q_{ch,aging}$ is used as a feature in this work (based on manual feature selection in Figure S5), the adaptive estimation algorithm is flexible because it allows the addition of multiple features for clustering along with the fine-tuning of other hyperparameters for better performance.

To conclude, variation in the initial SOH of SL batteries presents a significant challenge in accurately estimating the health of these batteries during their lifetime. Repurposed packs can contain cells with heterogeneities in their health, which means that as the battery is used, the cells go through different degradation trajectories, leading to safety concerns. BMS$_2$ is presented as a viable option that improves upon the existing health monitoring approaches to ensure SL batteries operate safely and for a long period of time. Through experimental data collected from eight retired pouch cells over a period of 15 months, it is observed that SL batteries, operated in a reduced voltage window of 3–4 V and HVAC-regulated temperatures, experience an increase in their C/20 capacity as the temperature increases. Battery capacity behavior is dominated by seasonal variation in temperature, with minimal signs of degradation from cycling. From the experimental data, 66 different features were extracted, and automatic feature selection was used to obtain a reduced number of features (six in total) for SOH estimation. Furthermore, five out of six features are such that they are

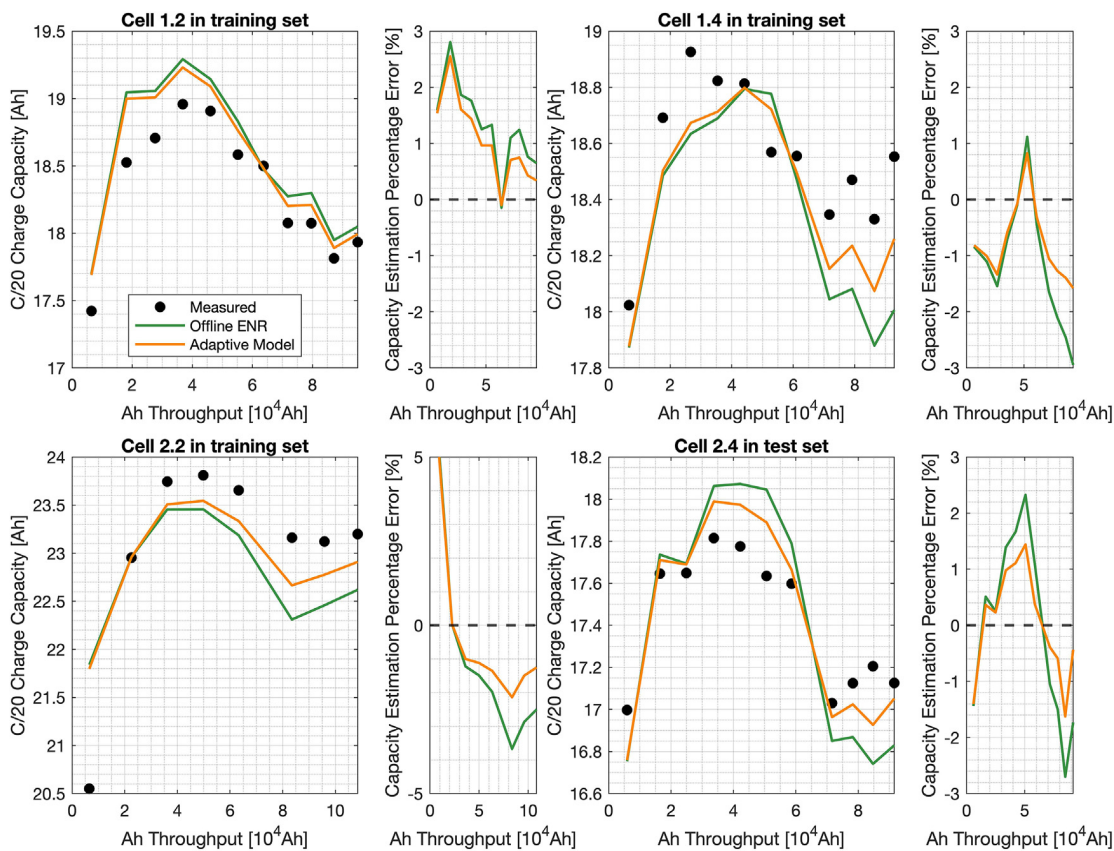

**Figure 6. Performance comparison between adaptive estimation model and offline ENR estimation model**

The x axis represents the Ah throughput, which monotonically increases with time. The adaptive model initially aligns with the ENR model and gradually converges to the true SOH. On both the training set and test set, the adaptive model outperforms the pure offline ENR estimation method.

accessible online for BMS₂ for real-time SOH estimation during battery operation. Four data-driven models, namely ENR, SVR, RFR, and GPR, were developed, and their performances were compared. Through significance testing, ENR was selected to be the best-performing model with a MAPE of less than 2.3%. Finally, an adaptive online SOH estimation algorithm was proposed that uses clustering between input features of training and test cells to bound the estimation error for onboard scenarios. The online SOH estimation model is capable of performing onboard BMS₂ for grid applications where temperature is not strictly controlled and current and voltage have uncertain variations based on the power demand throughout the year.

One limitation of this work is the size of the dataset. Longer battery testing times or testing more SL batteries would be helpful to verify and robustify the algorithms and consolidate the conclusions. Although the cardinality of the dataset is small, this work has revealed aspects of battery data, such as the inhomogeneity in testing temperatures, along with the non-monotonic capacity behaviors, that have not been documented in the literature, overall showing promise in the utilization of retired batteries. In terms of model choices, the limited size of the dataset prevents the use of big-data-based methods that are inherently better at generalizing to unseen data. In the future, it is hoped that the ENR model performance can be validated on new datasets from SL applications. Furthermore, the adaptive estimation algorithm can be extended to incorporate more features in the clustering step, and validated on larger datasets, ideally containing real field data.

## EXPERIMENTAL PROCEDURES

### Resources availability

*Lead contact*
Please contact Simona Onori (sonori@stanford.edu) for information related to the data and code described in the following experimental procedures section.

*Materials availability*
This study did not generate new unique reagents.

*Data and code availability*
Data and code are available at this link as supported by OSF (https://doi.org/10.17605/OSF.IO/FNS57).

### Automatic feature selection

To reduce the size of the feature set and identify features with a high correlation to SOH, automatic feature selection is performed using mRMR algorithm.[43] This algorithm has two parts: (1) maximum relevance objective $R$, which quantifies the mutual information between the input features and target output, and (2) minimum redundancy objective $S$, which quantifies the redundant features that are strongly correlated with one another. The overall objective of the mRMR algorithm is to maximize the difference between maximum relevance and minimum redundancy as $R - S$. The algorithm was implemented in MATLAB using the package `fsrmrmr`, and the resulting ranked features are shown in Figure 7. The top six features are selected to keep a balance between high-ranked features and the amount of features for the SOH estimation model since the features after the top six have a relatively flat importance score, which suggests that inclusion of these features will not significantly improve the estimation results.

### Extraction of selected features

Figure 8A shows one aging cycle with three current segments, $I_1$ and $I_2$ in discharge and $I_3$ in charge, and, similarly, three corresponding voltage segments, $V_1$, $V_2$, and $V_3$. Aging cycle accumulated Ah-throughput $Q_{Ah,aging}$ is extracted from the aging cycle current profile by integrating the absolute value of current with respect to time $t_0$ to $t_3$, as given in Equation 1. Similarly, Equation 2 gives the aging cycle charge energy-throughput $E_{ch,aging}$ by integrating the product of $I_3$ and $V_3$ with respect to time $t_2$ to $t_3$. This calculation is repeated for all aging cycles for each cell to get the complete trajectory of $Q_{Ah,aging}$ and $E_{ch,aging}$.

$$Q_{Ah,aging} = \int_{t_0}^{t_1} |I_1| dt + \int_{t_1}^{t_2} |I_2| dt + \int_{t_2}^{t_3} |I_3| dt \qquad \text{(Equation 1)}$$

$$E_{ch,aging} = \int_{t_2}^{t_3} |I_3 \cdot V_3| dt \qquad \text{(Equation 2)}$$

The two resistance features $R0_{ch,ch,low,2s}$ and $R0_{dis,ch,high,2s}$ are obtained from the charge portion of HPPC test as illustrated in Figure 8B. Based on Ohm's law, these resistance values are calculated using the sudden change in current and voltage; however, since these are 2 s charge-transfer (indicated by green dots) resistances, the voltage value also accounts for some relaxation voltage.

$$R0_{dis,ch,high,2s} = \frac{\Delta V_{R1}}{\Delta I_{R1}} \qquad \text{(Equation 3)}$$

$$R0_{ch,ch,low,2s} = \frac{\Delta V_{R2}}{\Delta I_{R2}} \qquad \text{(Equation 4)}$$

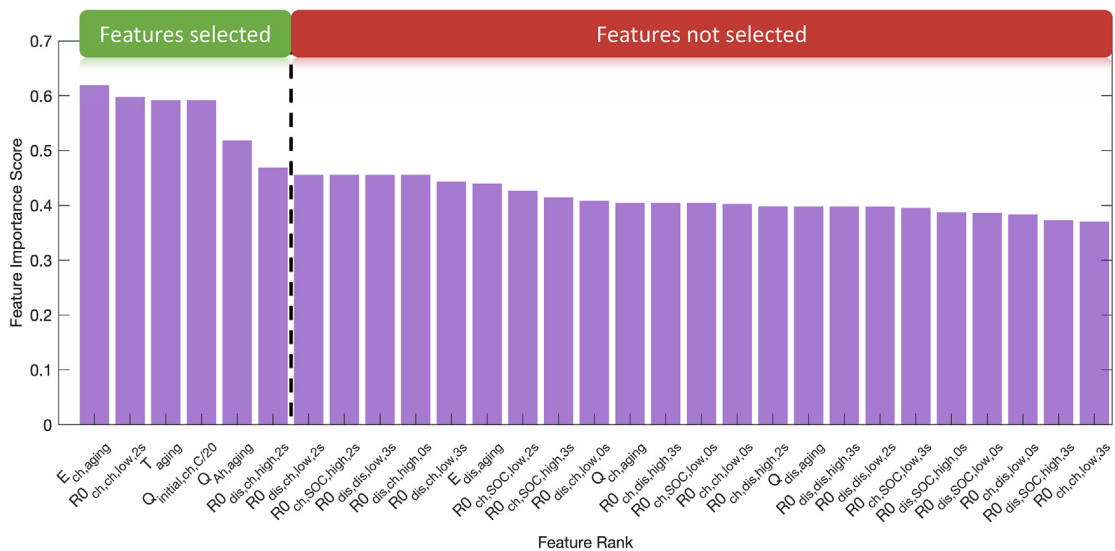

**Figure 7. Ranking and selection of features from mRMR algorithm**

On the training set (cells 1.1, 1.2, 1.3, 2.1, 2.2, and 2.3), the mRMR algorithm automatically ranks the features, and the top six features are selected, ranging from $E_{ch,aging}$ to $R0_{dis,ch,high,2s}$. The feature importance score for the remaining features stays relatively flat.

The temperature feature $T_{aging}$ is not explicitly calculated, but it is extracted by averaging the temperature during one aging cycle, and this process is repeated for all aging cycles for each cell.

**Performance metrics**

The performance of SOH estimation models is evaluated using three metrics: (1) RMSE, (2) RMSPE, and (3) MAPE. These are defined as follows:

$$RMSE(\hat{Y}, Y) = \sqrt{\frac{1}{M} \sum_{y \in Y, \hat{y} \in \hat{Y}} (\hat{y} - y)^2} \qquad \text{(Equation 5)}$$

$$RMSPE(\hat{Y}, Y) = \sqrt{\frac{1}{M} \sum_{y \in Y, \hat{y} \in \hat{Y}} \left(\frac{\hat{y} - y}{y}\right)^2} \times 100\%, \text{and} \qquad \text{(Equation 6)}$$

$$MAPE(\hat{Y}, Y) = \frac{1}{M} \sum_{y \in Y, \hat{y} \in \hat{Y}} \frac{|\hat{y} - y|}{y} \times 100\%, \qquad \text{(Equation 7)}$$

where $y \in Y$ represents the measured SOH, $\hat{y} \in \hat{Y}$ represents the estimated SOH from the model, and $M$ is the number of observations in the dataset.

**SUPPLEMENTAL INFORMATION**

Supplemental information can be found online at https://doi.org/10.1016/j.xcrp.2024.101941.

**ACKNOWLEDGMENTS**

Funding was provided by Stanford Storage X Initiative and Relyion Energy Inc. through the Stanford Storage X Initiative. Data used in this study were collected by Relyion Energy Inc. The authors would like to thank Stanford Energy Control Lab members Joseph Lucero, Yizhao Gao, and Maitri Uppaluri for proofreading the manuscript and testing the code.

**A** One aging cycle

**B** One charge HPPC

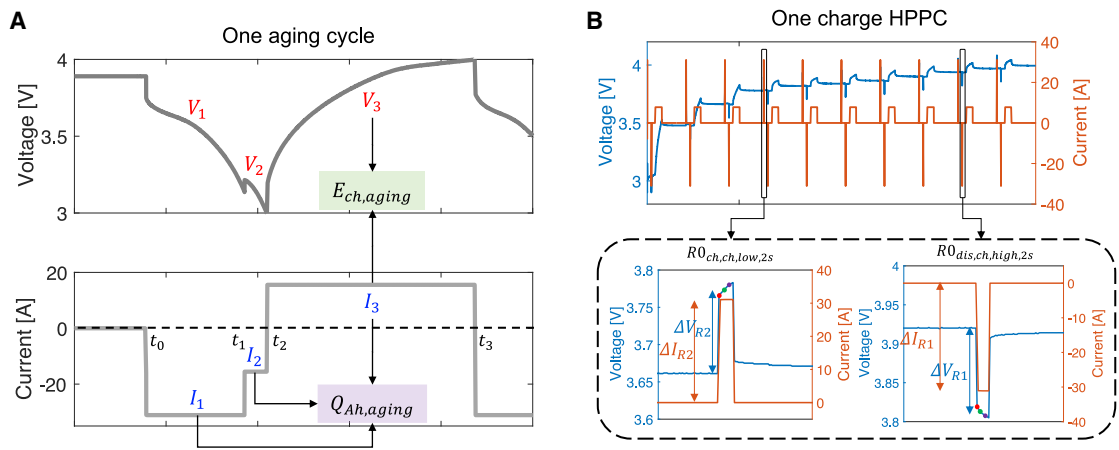

**Figure 8. Schematic of one aging cycle and one charge HPPC test**

(A) Voltage and current profiles are shown for a single aging cycle with three distinct current segments, $I_1$, $I_2$, and $I_3$, and three distinct voltage segments, $V_1$, $V_2$, and $V_3$. $Q_{Ah,aging}$ is calculated by integrating $I_1$, $I_2$, and $I_3$ with time, and $E_{ch,aging}$ is calculated by integrating the product of $I_3$ and $V_3$ with time.

(B) Charging portion of HPPC test between 3 and 4 V showing calculation of $R0_{ch,ch,low,2s}$ resistance (bottom left) and $R0_{dis,ch,high,2s}$ resistance (bottom right). The red, green, and purple dots on the voltage curve correspond to 0, 2, and 3 s charge-transfer resistances.

## AUTHOR CONTRIBUTIONS

Conceptualization, X.C., M.A.K., G.P., S.S., R.S., and S.O.; methodology, X.C., M.A.K., G.P., and S.O.; validation, X.C., M.A.K., and G.P.; formal analysis, X.C., M.A.K., G.P., and S.O.; investigation, X.C., M.A.K., G.P., S.S., R.S., and S.O.; data curation, S.S. and R.S.; writing – original draft, X.C., M.A.K., and S.O.; writing – review & editing (original draft), X.C., M.A.K., G.P., S.S., R.S., and S.O.; writing – review & editing (revised draft), X.C., M.A.K., and S.O.; visualization, X.C., M.A.K., and S.O.; supervision, project administration, and funding acquisition, S.O.; X.C. and M.A.K. contributed equally to this work, and their names are listed alphabetically based on the last names.

## DECLARATION OF INTERESTS

The authors declare no competing interests.